\documentclass[10pt,twocolumn,letterpaper]{article}

\usepackage[pagenumbers]{cvpr} 

\usepackage[dvipsnames]{xcolor}
\newcommand{\red}[1]{{\color{red}#1}}
\newcommand{\blue}[1]{{\color{blue}#1}}

\usepackage[accsupp]{axessibility}  
\usepackage{array}
\usepackage{multirow}
\usepackage{booktabs}
\usepackage{ulem} 

\usepackage{pifont}%
\newcommand{\cmark}{\textcolor{green}{\ding{51}}}%
\newcommand{\xmark}{\textcolor{red}{\ding{55}}}%

\definecolor{cvprblue}{rgb}{0.21,0.49,0.74}
\usepackage[pagebackref,breaklinks,colorlinks,citecolor=cvprblue]{hyperref}

\def\paperID{81} 
\def\confName{EarthVision}
\def\confYear{2024}

\title{GeoLLM-Engine: A Realistic Environment for Building Geospatial Copilots}

\author{Simranjit Singh, Michael Fore, Dimitrios Stamoulis\\
\textit{CoStrategist} R\&D Group, Microsoft Corporation, USA\\
{\tt\small \{simsingh, mifore, stamoulis.dimitrios\}@microsoft.com}
}

\begin{document}

\twocolumn[{
\maketitle\centering
\captionsetup{type=figure}
\includegraphics[width=0.99\textwidth]{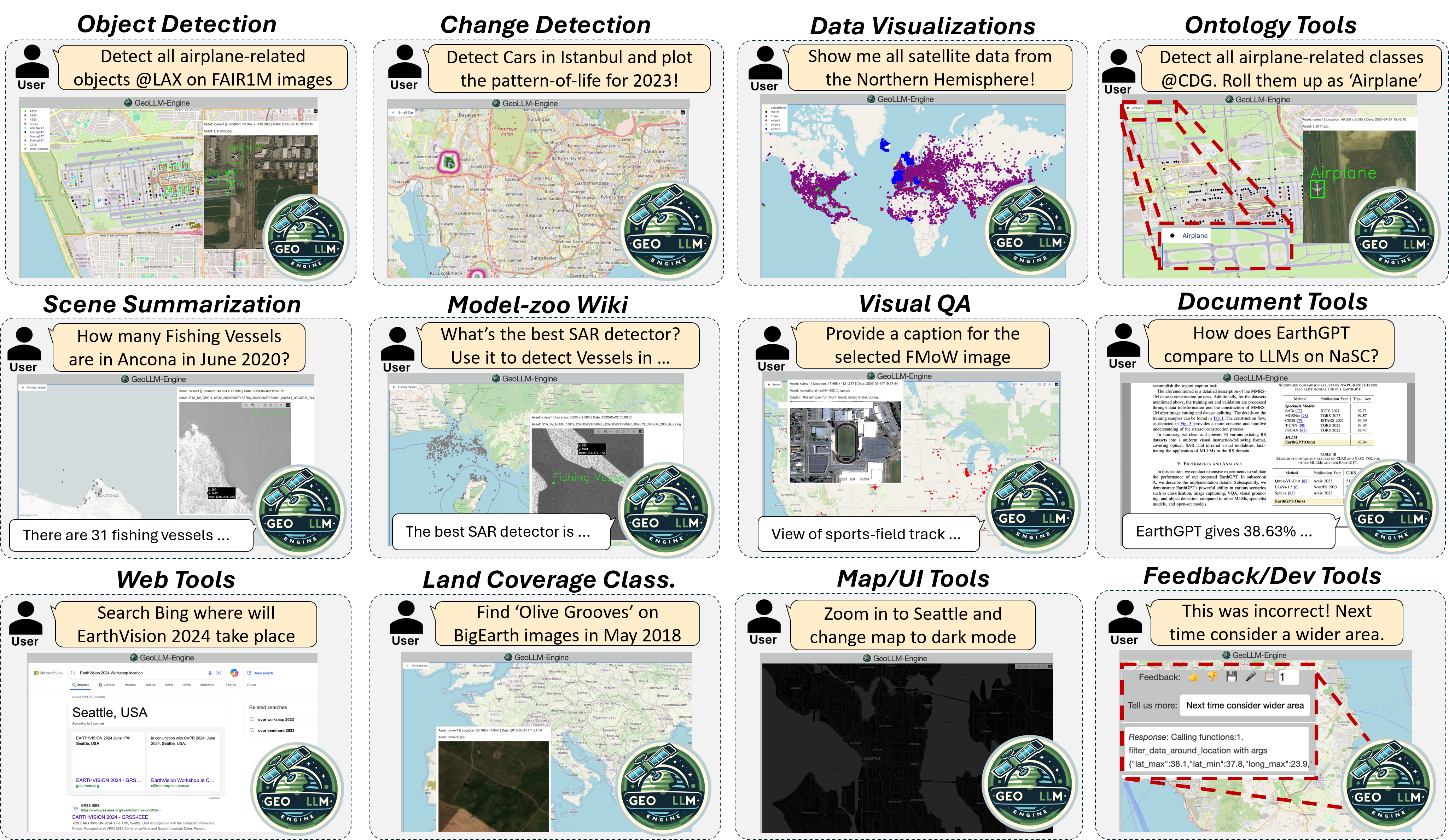}
\captionof{figure}{\texttt{GeoLLM-Engine} offers a \textit{realistic} web environment equipped with comprehensive APIs for developing, deploying, and evaluating geospatial Copilots on tasks that authentically reflect the workflows of remote sensing analysts. Unlike existing LLM benchmarks which rely on simplified question-answer pairs, \texttt{GeoLLM-Engine} introduces nuanced, long-horizon natural-language commands over a wide range of tasks - from object detection and land classification to document retrieval and web search.}
\label{fig:scenarios}\vspace{5mm}
}]

\maketitle

\begin{abstract}
Geospatial Copilots unlock unprecedented potential for performing Earth Observation (EO) applications through natural language instructions. However, existing agents rely on overly simplified single tasks and template-based prompts, creating a disconnect with real-world scenarios. In this work, we present \texttt{GeoLLM-Engine}, an environment for tool-augmented agents with intricate tasks routinely executed by analysts on remote sensing platforms. We enrich our environment with geospatial API tools, dynamic maps/UIs, and external multimodal knowledge bases to properly gauge an agent's proficiency in interpreting \textbf{realistic} high-level natural language commands and its \textbf{functional} correctness in task completions. By alleviating overheads typically associated with human-in-the-loop benchmark curation, we harness our massively parallel engine across 100 GPT-4-Turbo nodes, scaling to over half a million diverse multi-tool tasks and across 1.1 million satellite images. By moving beyond traditional single-task image-caption paradigms, we investigate state-of-the-art agents and prompting techniques against long-horizon prompts.
\end{abstract}    
\section{Introduction}
\label{sec:intro}

With the advent of generative AI, Large Language Models (LLMs) have the potential to significantly enhance Earth Observation (EO) workflows~\cite{wang2023skyscript, zhan2024skyeyegpt} across a broad range of tasks, from detection to learning from spatio-temporal data to analyzing aerial, UAV, and satellite images and videos~\cite{UN2024SDGs}. However, existing approaches consider predefined low-level template-based prompts that only capture the \textit{textual surface form} of the predicted image-caption pairs~\cite{zhou2023webarena}, while often overlooking the \textit{functional agent correctness} at completing high-level natural language (NL) commands~\cite{koh2024visualwebarena}. While the need for more representative benchmarks has been underscored across several generative AI domains~\cite{lecun2022path, maini2024tofu, zhou2023webarena}, their significance is even greater in the geospatial domain, as it involves complex multimodal data across diverse spatial and temporal dimensions.

Such disconnect partly stems from prevailing perceptions that benchmark creation, focused on simplistic single-task prompts, is straightforward. However, the overhead extends beyond merely curating benchmarks with image-caption pairs, a task that can be programmatically accomplished, as evidenced by the rapid release of numerous geospatial benchmarks over recent months~\cite{kuckreja2023geochat, zhang2024rs5m, guo2024remotesensinggpt, wang2023skyscript, zhan2024skyeyegpt, zhang2024earthgpt, roberts2024charting, yuan2024chatearthnet, mall2023remote, muhtar2024lhrsbot, singh2024geoqa}. Instead, the challenge lies in establishing an environment equipped with the requisite tools, dynamic UIs, and real-world APIs to form the ``engine'' for developing complex tasks. In this work, our \textbf{key insight} is that amidst the abundance of ``geospatial benchmarking'' works, a subtle refocus is necessary, prioritizing the construction of a robust \textit{engine as the foundation for benchmark creation}, rather than the benchmarks themselves. 

We draw inspiration from novel work~\cite{zhou2023webarena, maini2024tofu, zhuang2023toolqa} that introduces environment-based benchmarking suites for comprehensive agent assessment. While these works highlight potential in their domains, adopting them for geospatial applications necessitates overcoming human-in-the-loop bottlenecks, particularly in manual ground-truth verification and template creation. To mitigate these challenges, we employ formal-language-based verification techniques~\cite{yang2023finetuning}, recently introduced to expedite labor-intensive Reinforcement Learning from Human Feedback (RLHF) workflows. 

In this work, we introduce \texttt{GeoLLM-Engine}, a highly \textit{realistic} environment that captures real-world tasks on EO platforms~\cite{jakubik2023foundation}. Our environment comprises various fully operational APIs and dynamic map/web UIs to execute geospatial tasks via high-level NL prompts. More importantly, we employ model-correctness checker techniques~\cite{zhou2023webarena, yang2023finetuning} that allow our ``back-end'' engine to autonomously verify the accuracy of generated benchmarks, requiring only a one-off initial validation of task templates. By reducing the necessity for human intervention, we can massively parallelize our benchmark suite across 100 GPT-4-Turbo nodes to create large-scale benchmarks with 100,000 prompts that span half a million tasks over 1.1 million images from open-source EO datasets.

Each \texttt{GeoLLM-Engine} prompt exhibits high-level intent that emulates the nuances and abstract language usage patterns typically employed by human operators, as shown in \cref{fig:scenarios}. Using this benchmark, we follow state-of-the-art evaluation schemes with tool-augmented fine-tuning-free agents in zero-/few-shot in-context learning modality~\cite{koh2024visualwebarena}, powered by the latest GPT-3.5 and GPT-4 \texttt{Turbo (0125)} versions. Capturing our key insight, our findings reveal that merely expanding LLM benchmarks with more tasks of uniform complexity (\textit{e.g.}, an excess of visual QA captioning tasks) does not significantly enrich our understanding, as agent performance predictably shows little variation. Conversely, we show that varying levels of complexity better assess agent performance. To this end, we diversify the scope of our benchmark by incorporating various remote sensing (RS) applications over different satellite imagery sources and tasks of escalating intents, ranging from document knowledge retrieval to UI/Web interactions to geospatial data analytics.

\begin{figure}[t!]
\centering
\includegraphics[width=1.0\linewidth]{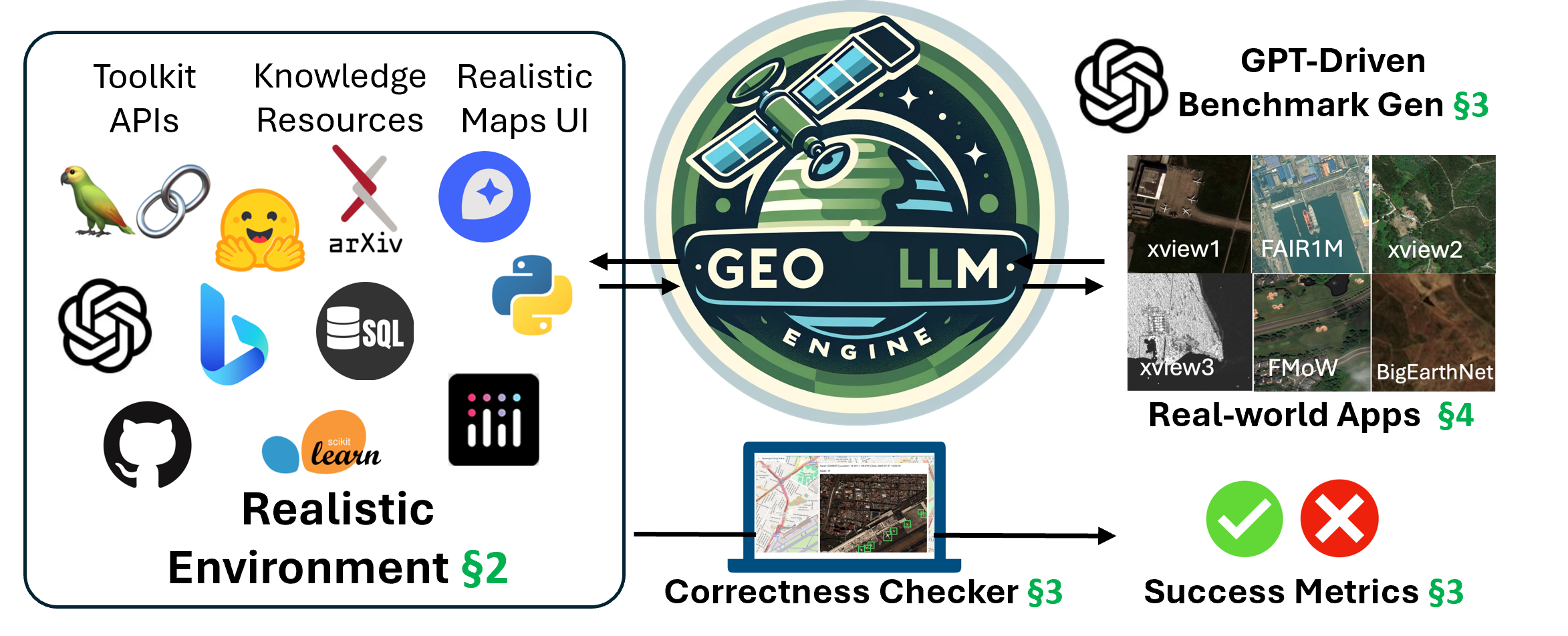}
\caption{\textbf{Paper overview:} \texttt{GeoLLM-Engine} is a realistic environment for building and evaluating agents on complex EO tasks. Our GPT-driven benchmark-\textit{engine} and the agent-correctness checker minimize the need for human intervention for benchmark curation. This allows us to develop a large-scale benchmark with more than $500,000$ \textit{long-horizon} geospatial tasks on $1,100,000$ satellite images.}
\label{fig:geo_engine}
\end{figure}

\section{\texttt{GeoLLM-Engine} Environment}
\label{sec:methodology}

Our aim is to establish a realistic environment that clearly advances beyond current benchmarks, featuring a self-contained web UI with a varied array of LLM tools and an integrated benchmarking engine (\cref{fig:geo_engine}). Built upon open-source libraries and APIs, \texttt{GeoLLM-Engine} is designed as a reproducible and scalable platform to support the development and evaluation of geospatial agents.

\textbf{Environment - ``Front-end'':}
A key challenge in creating such an environment is the need for reproducibility and comparability across different systems and methodologies. To address this, we leverage a suite of open-source APIs, enabling the seamless integration of a wide array of tools, datasets, and functionalities, while facilitating transparency and accessibility. We intend to release our codebase and benchmark to foster advancements in geospatial Copilots.

\begin{table}[t!]
  \centering
  \resizebox{\columnwidth}{!}{%
  \begin{tabular}{@{}lccc@{}}
    \toprule
    Tool Types & \# Tools & Tool Examples \\
    \midrule
    \includegraphics[width=0.4cm,keepaspectratio]{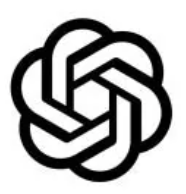} OpenAI & 5 & \texttt{json\_mode()}, \texttt{embeddings()}  \\
    \includegraphics[width=0.4cm,keepaspectratio]{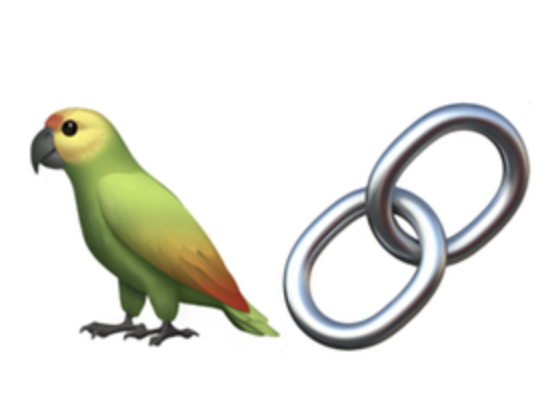} LangChain & 9 & \texttt{vectorstores.FAISS()}  \\
    \includegraphics[width=0.4cm,keepaspectratio]{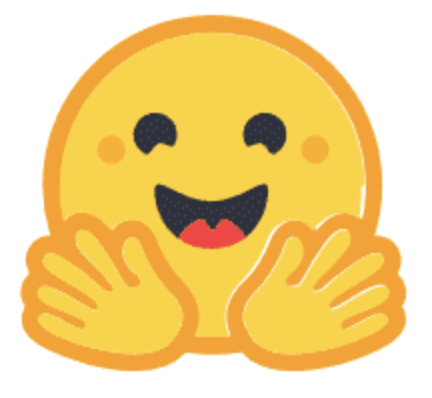} Vision & 17   & \texttt{transformers.SwinModel()} \\
    \includegraphics[width=0.4cm,keepaspectratio]{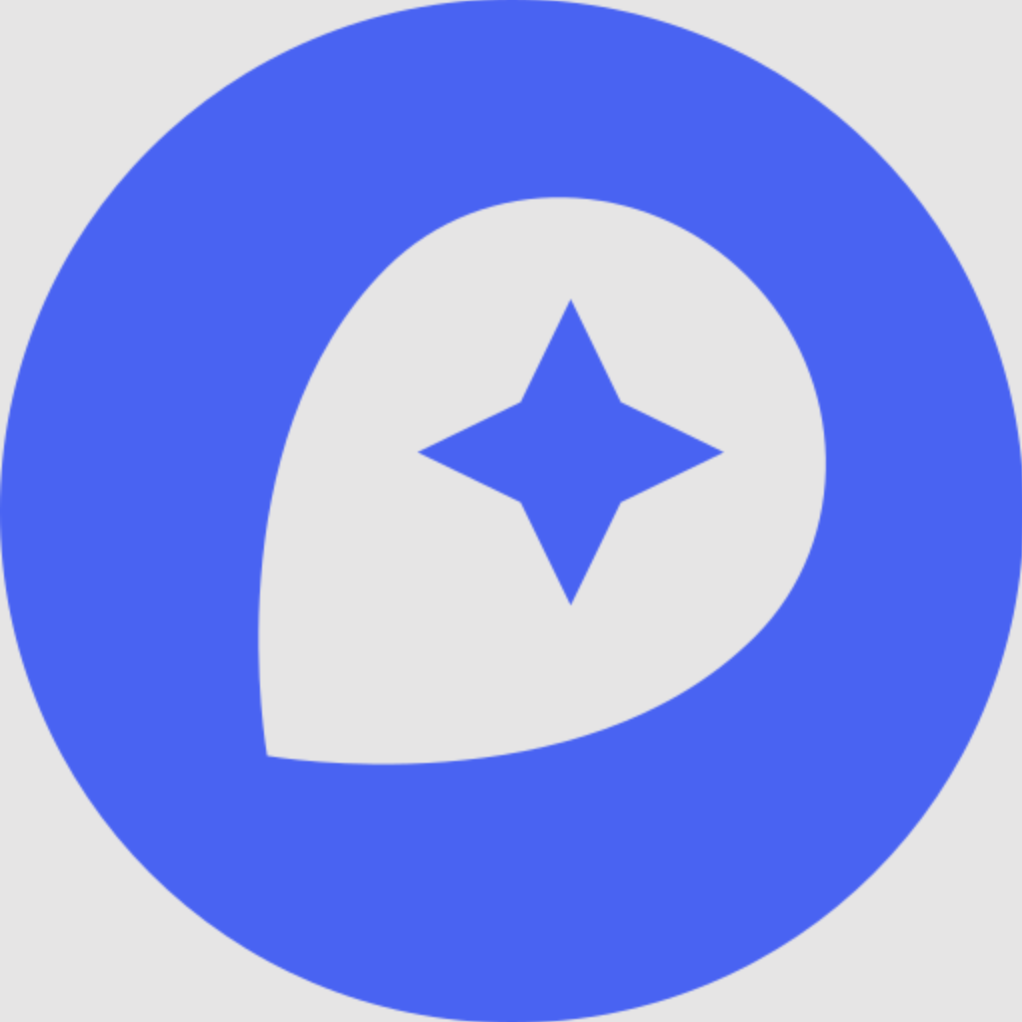} Map & 31  & \texttt{scattermapbox()} \\
    \includegraphics[width=0.4cm,keepaspectratio]{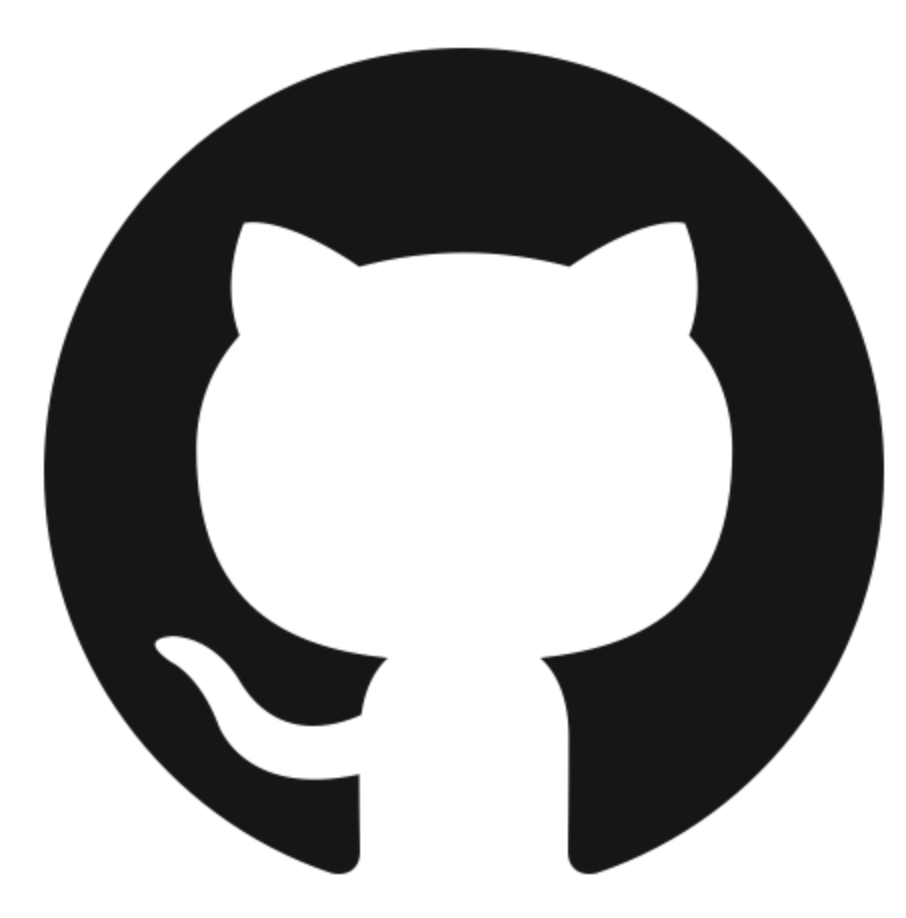} Geotools & 23  & \texttt{rasterio()}, \texttt{geopandas()} \\
    \includegraphics[width=0.4cm,keepaspectratio]{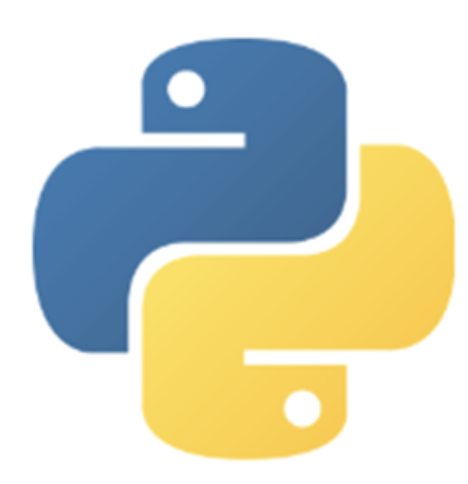} Python & 20   & \texttt{pyPDF()}, \texttt{PIL()} \\
    \includegraphics[width=0.4cm,keepaspectratio]{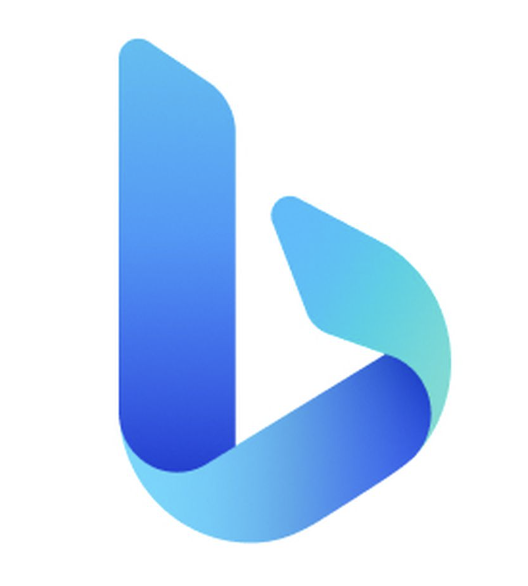} Web & 15  & \texttt{openURL()}, \texttt{bing\_search()}  \\
    \includegraphics[width=0.4cm,keepaspectratio]{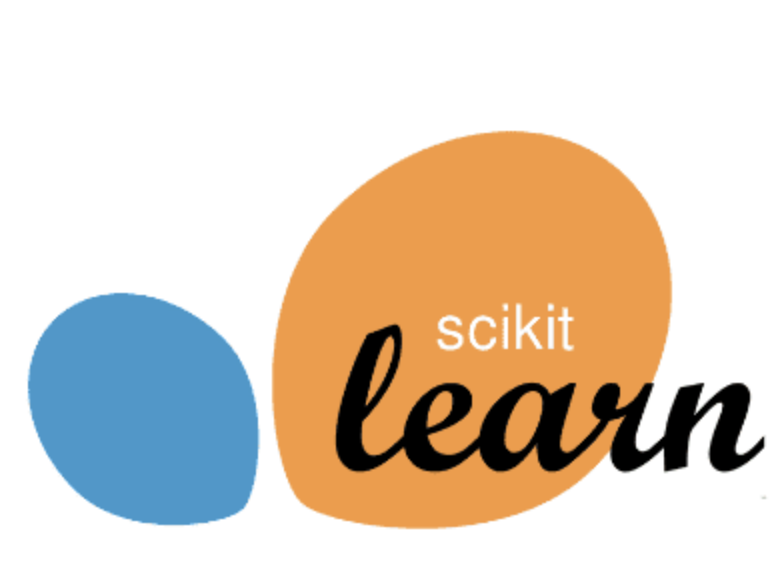} Data Tools & 28  & \texttt{sklearn.cluster()}  \\
    \includegraphics[width=0.4cm,keepaspectratio]{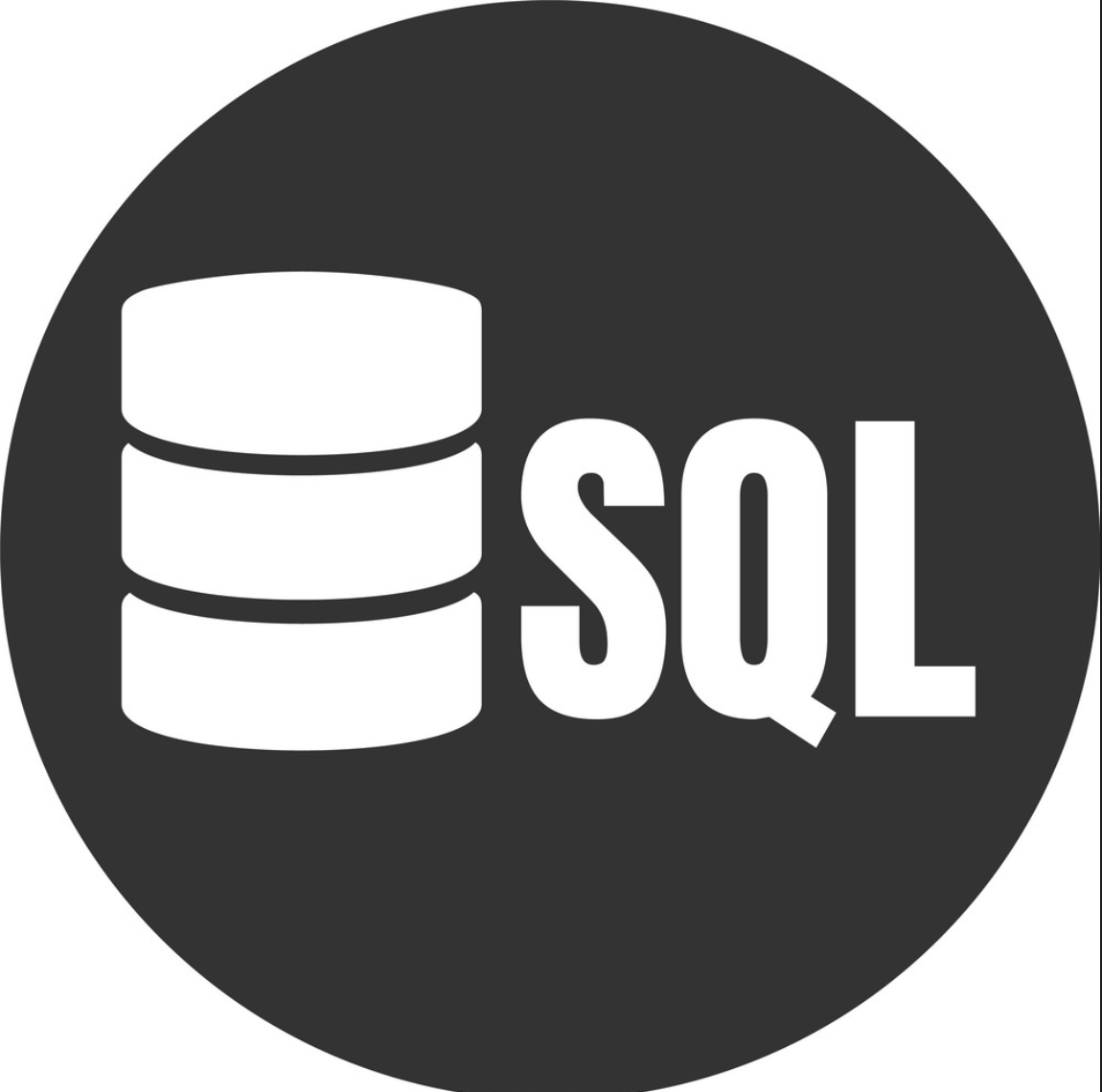} Database & 8   &  \texttt{db\_load()}, \texttt{filter\_date()}  \\
    \includegraphics[width=0.4cm,keepaspectratio]{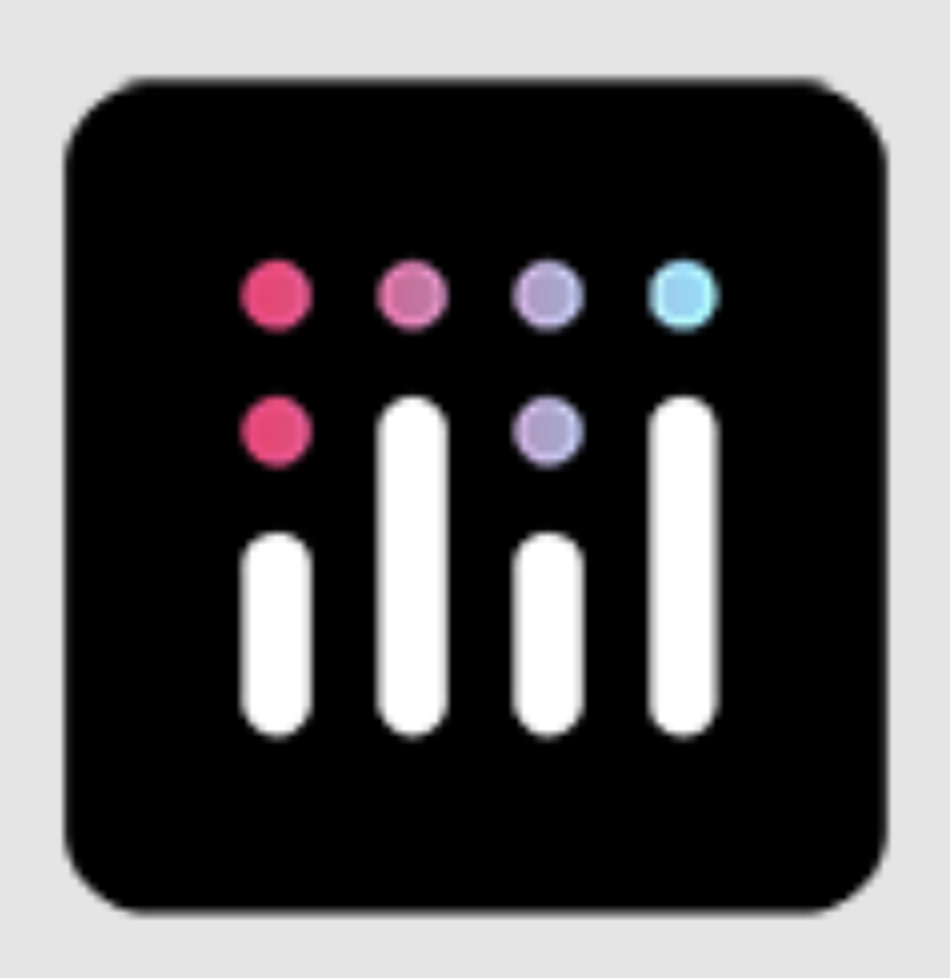} UI & 12  & \texttt{display\_hover()} \\
    \includegraphics[width=0.4cm,keepaspectratio]{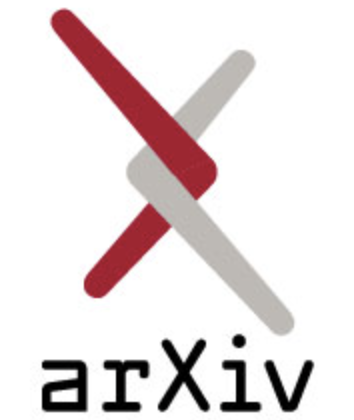} Knowledge & 6  & \texttt{wiki\_RAG()}, \texttt{model\_cards()} \\
    \bottomrule
  \end{tabular}
  }
  \caption{\texttt{GeoLLM-Engine} module inventory: Tool Space $\mathcal{T}$  }
  \label{tab:tools_inventory}
\end{table}

\textbf{Tool Space:}
We equip our environment with a comprehensive array of open-source Python packages, catering to various functionalities from data analytics to LLM-specific tasks, such as employing \texttt{LangChain} for FAISS~\cite{douze2024faiss} embeddings in knowledge retrieval applications. The interface leverages \texttt{Mapbox} APIs for interactive mapping, while \texttt{rasterio} and \texttt{geopandas} facilitate advanced manipulation of geospatial data. The complete module inventory, comprising over 175 tools, is presented in \cref{tab:tools_inventory}. This diverse toolkit enables us to execute complex tasks ranging from satellite imagery analysis to utilizing vector storage for rapid geographic data querying. 

\textbf{Engine - ``Back-end'':}
To facilitate large-scale experimentation, \texttt{GeoLLM-Engine} incorporates a Command-Line Interface (CLI) alongside its UI, providing comprehensive tooling and scripting capabilities. This feature is crucial for conducting extensive investigations, allowing us to efficiently run model-verification checks on over half a million tasks and evaluate baseline agents against our benchmarks. Thanks to this setup, these operations are completed within hours, leveraging hundreds of GPT endpoints. The CLI, in tandem with the UI, provides diverse modalities essential for replicating the intricate demands of geospatial analysis tasks which require the integration of disparate data sources and analytical techniques.

\textbf{Environment Formulation:}
Our environment is represented as $\mathcal{E}=\langle \mathcal{S}, \mathcal{A}, \mathcal{T} \rangle$, comprising the environment state $\mathcal{S}$, the action space $\mathcal{A}$, and the tool space $\mathcal{T}$ (\cref{tab:tools_inventory}). To intuitively understand $\mathcal{S}$, consider a user query ``\textit{zoom into the Indo-Pacific region and show me the vessels during November 2021}.'' Upon query completion, $\mathcal{S}$ will encapsulate not just the visual or textual response, but also the altered state of the map, such as its zoom level or position, the loaded database, and the temporal window of displayed data.

While detailed notation of transition functions extends beyond our study's scope, previous research~\cite{zhou2023webarena} has established these functions as deterministic. This implies a critical property: given the same starting state $\mathcal{S}_0$, any two agents executing an identical sequence of actions or tools will invariably arrive at the same final state $\mathcal{S}^*$. This deterministic nature is vital for our purposes, as it allows for the verifiability of any agent's solution against the benchmark's ``ground truth'' specifications~\cite{yang2023finetuning}. We exploit this feature in our model checker (elaborated in \cref{sec:benchmark_suite}) in two principal ways. Firstly, given a ``golden'' ground truth, we can deterministically assess the functional correctness of any candidate solution that follows the same sequence of actions or tools. Secondly, this principle underpins our approach to benchmark (ground-truth) generation: by evaluating multiple agent solutions, a consensus on the final state among the majority indicates a high likelihood of an accurate ground truth, hence eliminating the need for human inspection.

\textbf{User Intent Formulation:}
Intuitively, each user intent is encapsulated by four parts: the question $q$ that triggers the agent, the executed tool sequence $\hat{T}$, the agent's textual response $\hat{r}$ to the user, and the concluding environment state $\hat{S}$. Thus, we can express each task as $\{q, \hat{T}, \hat{r}, \hat{S}\}$. The sequence $\hat{T}$ is defined by the set of tool $\hat{T} = \{\hat{t}_1, \hat{t}_2, \dots\}$, where at each step $i$ the agent invokes tool $\hat{t}_i= \{\hat{tool}_i, \hat{args}_i\} \in \mathcal{T}$. As shown later in \cref{sec:benchmark_suite}, by contrasting the task set $\{q, \hat{T}, \hat{r}, \hat{S}\}$ with a gold standard $\{q, T^*, r^*, S^*\}$, we can ascertain the functional correctness across our entire benchmark.

\section{\texttt{GeoLLM-Engine} Benchmark Suite}
\label{sec:benchmark_suite}

\begin{table*}[ht]
  \centering
  \resizebox{\textwidth}{!}{%
  \begin{tabular}{@{}lcc@{}}
    \toprule
    Category & Example & Tools $T$ \\
    \midrule
    \multirow{4}{*}{Load$\rightarrow$Filter$\rightarrow$Plot} & Plot on the map the fair1m and xview1 images around Pittsburgh, PA, USA  & \texttt{db\_load()}, \texttt{filter\_coords()}, \\
    & from May 2023 for all vehicle-related categories & \texttt{filter\_date()}, \texttt{filter\_class()}, $\dots$ \\\cline{2-3}
    & \multirow{2}{*}{Show me all \textit{golf course} LCC images for FMoW from January 2015 to May 2016}  & \texttt{db\_load()}, \texttt{filter\_lccclass()}, \\
    & & \texttt{filter\_date()}, \texttt{scatter\_plot()}, $\dots$ \\\cline{2-3}
    & Fetch all Fishing-Vessel detections on xview3 from Belle \^Ile en Mer, France & \texttt{db\_load()}, \texttt{filter\_coords()}, $\dots$ \\
    \midrule
    \multirow{2}{*}{UI/Web Navigation} & Can you search Google for ``Foundation models in Earth Observation apps''? & \texttt{google\_search()}, \texttt{open\_url()} \\\cline{2-3}
    & Can you zoom the map to Seattle, WA? & \texttt{zoom\_map()} \\
    \midrule
    \multirow{3}{*}{Information Seeking} & How many distinct semantic tags does SkyScript (arxiv) covers and what & \texttt{arxiv\_search()}, \texttt{docs\_RAG()},  \\
    & percentage of image-tag pairs was \textit{manually} checked for tag accuracy? &  \texttt{answer\_tools()} \\\cline{2-3}
    & According to our Model wiki, which detector is best-suited for vessels on SAR imagery? &  \texttt{modelzoo\_search()}, \texttt{answer\_tools()} \\
    \bottomrule
  \end{tabular}
  }
  \caption{Examples of \textit{realistic} \texttt{GeoLLM-Engine} tasks across three primary ``Intent'' categories. Please note the \textit{long-horizon high-level} natural language commands, which constitutes a novel departure from existing single-task template-based geospatial LLM benchmarks.}
  \label{tab:user_queries}
\end{table*}

In this section, we first describe the process for ``\textit{grounding}'' high-level natural language instructions into a structured set of task templates and user intents that cover all the \texttt{GeoLLM-Engine} tools. Next, we discuss our GPT-driven, human-out-of-the-loop  ground-truth sampling approach.

\textbf{Intent Collection:} 
The initial phase involves utilizing human annotators (our team members) to craft a small set of user intents, following the qualitative guidelines set forth in~\cite{zhou2023webarena}. These intents are designed to be both nuanced and high-level, requiring the agent to perform more than one or two actions, and should be decomposable into a series of interchangeable templates. While this step is manual, please note that it represents the primary (and essentially the only) offline step necessitating human involvement. Through this procedure, we generate the foundational intents and templates, serving as the modular components from which benchmark queries are constructed.

The annotators are instructed to input detailed prompts, utilizing zero-shot GPT-4-Turbo with Chain-of-Thought prompting~\cite{wei2023chainofthought} to propose solutions (as highlighted in our Results, GPT-4 demonstrates notable zero-shot capabilities). Utilizing ``User Feedback'' UI buttons, annotators identify and confirm correctly executed model responses, hence collecting the ``correct'' examples along with the corresponding agent actions. By promoting scenarios that necessitate an average of 7-8 tool interactions, we can cover the entire tool space within 250 instantiated queries. Following the granular intent categories as in~\cite{zhou2023webarena}, we classify the various prompts into the categories shown in \cref{tab:user_queries}:

\begin{enumerate}
    \item \textbf{Information Seeking}: Queries for knowledge retrieval aimed at sourcing information from wikis and documents to support EO investigations.
    \item \textbf{UI/Web Navigation}: Commands designed for UI interaction, such as opening web search results or displaying images corresponding to detections mapped out.
    \item \textbf{Load-Filter-Plot}: Operations to load data, apply specific filters, and present geospatial findings in an insightful manner, for example, through change-detection heatmaps or land cover classification (LCC) categories.
\end{enumerate}

\begin{table*}[ht!]
  \centering
  \resizebox{\textwidth}{!}{%
  \begin{tabular}{@{}lcccc@{}}
    \toprule
    Model Checks & Function & Description & Condition & Check Failed? \\
    \midrule
    \textbf{\textit{Correctness}} & & \\
    \multirow{3}{*}{Tool Calls} & \blue{$f_{valid}(\hat{tool})$} & Agent selected ``real'' tool & $tool \in $ tool space $\mathcal{T}$ & \red{Infeasible Action} \\
     & \blue{$f_{needed}(\hat{tool})$} & Tool was required to solve the task & $\hat{tool} \in $ ground-truth tools $T$ & \red{Function Error} \\
     & \blue{$f_{missed}(tool^*)$} & Agent missed the tool from its solution & $tool^* \in $ $T || tool^* \notin $ solution $\hat{T}$ & \red{Missed Function} \\[1ex]\cline{2-5}
    \multicolumn{2}{c}{} \\[-1ex]
    \multirow{2}{*}{Tool Args} & \blue{$f_{str}(\hat{args}, args^*)$} & Input variables (string) names matching & $\forall arg_i: \hat{arg}_i == arg^*_i $ & \multirow{2}{*}{\red{Argument Error}} \\
     & \blue{$f_{num}(\hat{args}, args^*)$} & Input variables (float) ``matching'' & $\forall arg_i: |\hat{arg}_i - arg^*_i| < \epsilon $  & \\
    \midrule
    \textbf{\textit{Success}} & & \\
    \multirow{2}{*}{Agent Output} & \blue{$f_{reply}(\hat{r}, r^*)$} & Response reasonably close to expected answer & $ROUGE(\hat{r}, r^*) > \epsilon_R$ & \multirow{4}{*}{\red{Unsuccessful Task}} \\
     & \blue{$f_{output}(\hat{o}, o^*)$} & Correct output (\textit{e.g.}, detected objects) &  &  \\[1ex]\cline{2-4}
    \multicolumn{2}{c}{} \\[-1ex]
    \multirow{2}{*}{Tool Args} & \blue{$f_{map}(\hat{S}_{map}, S_{map}^*)$} & Correct map view (\textit{e.g.}, zoomed area) & $\hat{S}_{map} == S_{map}^*$ &  \\
     & \blue{$f_{map}(\hat{S}_{UI}, S_{UI}^*)$} & Correct UI view (\textit{e.g.}, open web-tabs) & $\hat{S}_{UI} == S_{UI}^*$ &   \\
    \bottomrule
  \end{tabular}
  }
  \caption{We define \textit{model-checks} to obtain the agent success and correctness at performing each task.}
  \label{tab:checker}
\end{table*}

\textbf{Tool Templates:}
Given the 250 instantiated queries and GPT's solutions, we programmatically parse all \texttt{json} responses and remove duplicates, obtaining a \textbf{distinct template for every tool agent-call}, \textit{i.e.}, $\forall t_i \in \mathcal{T}$ we have the GPT \texttt{json}-call $\{toolname_i, args_i\}$. These \texttt{json} responses furnish us with the ``building blocks'' that can be used for benchmark sampling in a straightforward intuition: leveraging the recently introduced \texttt{json\_mode} feature in OpenAI's APIs, we can initiate queries to a standalone GPT model. Within this context, we present the model with a specific tool interaction template and pose inquiries akin to: ``\textit{Given your prior action of zooming into X and the corresponding function call, generate the new function call to instead focus on Vienna}.'' This allows us to dynamically generate tool-specific commands tailored to new, contextually relevant scenarios.

\textbf{GPT-Driven Benchmark Creation:}
To efficiently scale our benchmark generation, we capitalize on two key attributes of GPT-4-Turbo agents: the extended input token lengths and their ability to parse extensive databases described through SQL-like schemas. We leverage the expanded token capacity and we incorporate directly into the model's context \textbf{all} tool-calling \texttt{json} templates for the 174 tools (manually verified in the previous step). Moreover, we compile the metadata (\textit{e.g.}, coordinates, dates, categories, document titles, \textit{etc}.) associated with all satellite images and documents into a SQL table. From this, we randomly select 1,000 entries, collecting their categories, coordinates, and dates sets and providing them as SQL schemas to GPT. Last, we append the 250 manually crafted queries as ``successfully sampled'' examples, and alongside in-context ``benchmark creation'' instructions, this entire prompt is fed to GPT-4, prompting it to autonomously generate both a suggested task prompt and the corresponding solution, forming a new query task $\{q, T^*, r^*, S^*\}$.

To circumvent the need for manual verification of each solution's accuracy, we introduce a novel approach that integrates functional model checking with the principle of LLM \textit{self-consistency}~\cite{wang2023selfconsistency}. Drawing from the concept where an agent repeatedly solves the same prompt, typically converging on the correct solution, we apply this by having GPT-4 propose multiple solutions to its self-generated (original) prompt. After 10 iterations, we execute these solutions in our engine (in-parallel via our CLI tools) to ascertain the final state $S^*$ for each. If 9 out of 10 solutions converge on the same end state, we consider this to be a verified ground truth and incorporate it into our dataset. Overall, this novel \textit{self-consistency} sampling scheme allows us to streamline the benchmark generation process without extensive human intervention.

\textbf{Model-Checker Formulation:}
\texttt{GeoLLM-Engine} incorporates a rigorous set of model checks to determine both the functional correctness and overall success of an agent’s response to a given prompt. As shown in \cref{tab:user_queries}, \textit{correctness} checks the agent's tool usage $\hat{T}$ against the corresponding ground-truth $T^*$, assessing argument accuracy within function calls (\textit{e.g.}, missing a required tool or calling the right tool with wrong parameters). \textit{Success} checks the final system state produced by the agent's sequence of actions $\hat{S}$ against the expected ground-truth state $S^*$. We leverage our engine to ``run'' both the sequences starting from the same initial state to confirm whether the agent's output and ground-truth match. Note the distinction between the two failure cases; they are not mutually inclusive. For example, an agent may erroneously invoke an unnecessary tool (``Function Error''), yet this may not alter the final state, which could still align with the anticipated result.

\textbf{Agent Evaluation Metrics:}
Based on the model checks, we can define the appropriate metrics that we subsequently use to evaluate the performance of different agents: 
\begin{enumerate}
    \item \textit{Success rate}: the ratio of successfully completed tasks across the entire benchmark as defined by $f_{success}$ (\cref{tab:checker}). This ratio informs us of the degree to which the agent is able to complete tasks, irrespective of whether it took incorrect or unnecessary intermediate steps. 
    \item \textit{Correctness rate}: the ratio of correct function-call operations across the benchmark as defined by the $f_{correctness}$ error types. Given the total number of errors and ground-truth tools, we compute the correctness ratio $R_{correct} = \max(0, 1- N_{errors}/N_{tools})$~\citep{zhuang2023toolqa, maini2024tofu}, which captures how likely it is for the agent to invoke the correct functions in the expected order.
    \item \textit{ROUGE score}: ROUGE-L recall score~\citep{lin04rouge} to compare final model replies $\hat{r}$ with the ground truth $r^*$.
\end{enumerate}

\begin{figure*}[ht!]
\centering
\includegraphics[width=1.0\linewidth]{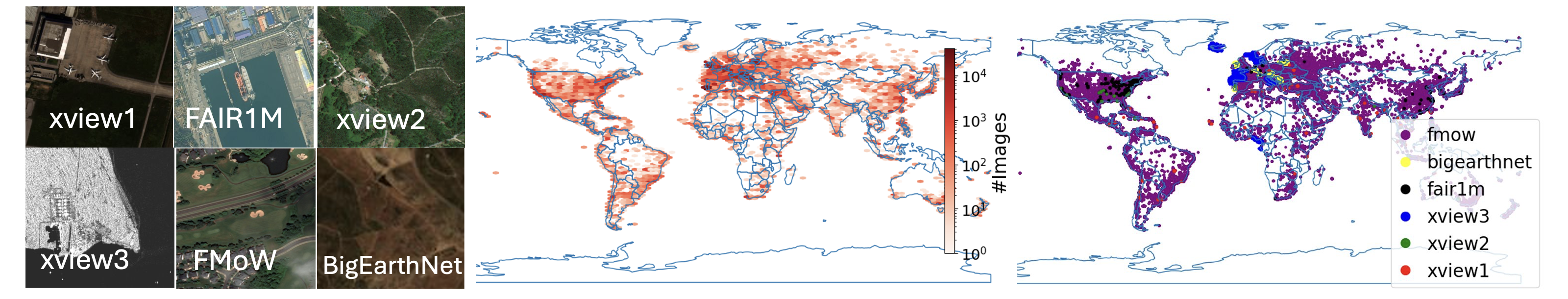}
\caption{Dataset overview with a wide spatial and temporal range. The satellite imagery serves as \textit{task contexts} for LLM agents to perform function calls (\textit{e.g.}, given image coordinates, we can assess the agents' ability to fetch the proper set of images around a user-specified location) and is not employed for any LLM-finetuning or other downstream tasks, enabling our research-purposes investigation.}
\label{fig:datasets}
\end{figure*}

\begin{table*}[ht!]
  \centering
  \resizebox{0.9\textwidth}{!}{%
  \begin{tabular}{@{}lccccccc@{}}
    \toprule
    \multirow{2}{*}{Method} & Correctness & Success & Obj. Det & LCC  & VQA & Avg. Tokens \\
    ~      & Rate $\uparrow$  & Rate $\uparrow$  & F1 $\uparrow$  & R $\uparrow$  & Rouge-L $\uparrow$  & / Task $\downarrow$ \\
    \midrule
    \textbf{\textit{GPT-3.5 Turbo (0125)}} & & & & & & & \\
    Chameleon~\cite{lu2023chameleon} Zero-Shot & 38.08\% & 40.94\% & 39.18\% & 49.52\% & 52.74\% & 38.9k\\
    Chameleon~\cite{lu2023chameleon} Few-Shot & 60.31\% & 48.31\% & 43.85\% & 44.29\% & 54.62\% & 39.7k \\
    CoT~\cite{wei2023chainofthought} Zero-Shot & 40.20\% & 64.66\% & 54.99\% & 93.28\% & 54.40\% & 21.1k\\
    CoT~\cite{wei2023chainofthought} Few-Shot & 65.77\% & 68.45\% & 73.81\% & \textbf{98.39\%} & 56.28\% & 26.4k\\
    ReAct~\cite{yao2023react} Zero-Shot & 62.06\% & 51.56\% & 54.16\% & 92.57\% & 56.45\% & 17.3k\\
    ReAct~\cite{yao2023react} Few-Shot & \textbf{68.42\%} & \textbf{73.47\%} & \textbf{75.01\%} & 97.45\% & \textbf{65.26\%} & 30.9k\\
    \midrule
    \textbf{\textit{GPT-4 Turbo (0125)}} & & & & & & & \\
    Chameleon~\cite{lu2023chameleon} Zero-Shot  & 46.31\% & 57.10\% & 45.57\% & 91.20\% & 49.76\% & 38.3k\\
    Chameleon~\cite{lu2023chameleon} Few-Shot & 67.78\% & 59.01\% & 55.82\% & 67.92\% & 49.27\% & 40.7k\\
    CoT~\cite{wei2023chainofthought} Zero-Shot & 80.88\% & 77.35\% & 87.99\% & 96.56\% & 65.29\% & 23.6k\\
    CoT~\cite{wei2023chainofthought} Few-Shot & 84.01\% & 80.00\% & 88.40\% & \textbf{99.89\%} & 67.65\% & 25.8k\\
    ReAct~\cite{yao2023react} Zero-Shot & 84.27\% & 80.03\% & \textbf{89.34\%} & 98.83\% & 68.11\% & 26.7k\\
    ReAct~\cite{yao2023react} Few-Shot & \textbf{84.31\%} & \textbf{81.11\%} & 83.85\% & 99.63\% & \textbf{69.37\%} & 32.5k\\
    \bottomrule
  \end{tabular}
  }
  \caption{The overall performance of different methods and prompting strategies on \texttt{GeoLLM-Engine-10k}.}
  \label{tab:results_summary}
\end{table*}

\section{Remote Sensing Datasets}
\label{sec:datasets}

We consider several open-source data sources encompassing tasks like object detection, land cover classification, and visual question answering to build a collective dataset of \textbf{1,149,612} images. All data sources have coordinates and time metadata which offer us global spatio-temporal coverage in our benchmark prompts (\cref{fig:datasets}): 
\begin{enumerate}
    \item \textit{xView1}~\cite{lam2018xview}: 846 high-resolution images sourced from WorldView-3 satellites focusing on overhead object detection with 1 million objects and 60 classes. 
    \item \textit{xView2}~\cite{gupta2019xbd}: dataset for building damage assessment with 5,598 images sourced from the Maxar/DigitalGlobe Open Data Program before and after 19 natural disasters with 850,736 building annotations. 
    \item \textit{xView3}~\cite{paolo2022xview3sar}: a dataset of 23,432 SAR GRD images from the Sentinel-1 mission annotated for (fishing) vessel detections to study illegal fishing practices. 
    \item \textit{SARFish}~\cite{Luckett_2024_WACV}: extends the xView3-SAR GRD dataset by providing products from the Sentinel-1 C-band SAR satellite constellation operated by the European Space Agency’s (ESA) Copernicus Program in both real-valued GRD and complex-valued SLC product types. 
    \item \textit{FAIR1M}~\cite{sun2022fair1m}: object detection dataset with 24,775 images with over 1 million instances sourced from Gaofen satellites and Google Earth.  
    \item \textit{Functional Map of the World (FMoW)}~\cite{christie2018functional}: a multi-label LCC dataset for buildings/land use with 727,144 images from over 200 countries. 
    \item \textit{BigEarthNet}~\cite{sumbul2019bigearthnet}: 344,385 Sentinel-2 LCC images from the CORINE Land Cover database.
\end{enumerate}
\section{Results}
\label{sec:results}

\cref{tab:results_summary} summarizes the performance of state-of-the-art GPT-based agents with various prompting schemes. In addition to the LLM metrics, we report the agent's performance with respect to the underlying object detection (F1 score), LCC (Recall), and visual question answering (VQA, Rouge-L) tasks in our benchmark. These metrics provide insights into each agent's efficiency, accuracy, and responsiveness in executing geospatial tasks.

\textbf{GPT versions:}
We observe that there's a marked difference in performance between models based on the GPT-3.5 Turbo and GPT-4 Turbo frameworks, with the latter generally achieving higher scores across all metrics. For instance, both the Chain-of-Thought~\cite{wei2023chainofthought} (CoT) and ReAct~\cite{yao2023react} methods, especially in their few-shot configurations, demonstrate a significant leap in correctness and success rates when transitioning from GPT-3.5 to GPT-4. This improvement underscores the advancements in model understanding and task execution capabilities. This finding is consistent with most of the work in language guided agents~\cite{bubeck2023sparks}. As anticipated, the few-shot configurations outperform their zero-shot counterparts across all evaluated methods. This trend underscores the value of providing models with a few examples to adapt to specific tasks, significantly enhancing their ability to accurately interpret and respond to complex geospatial queries.

\begin{figure*}[t!]
\centering
\includegraphics[width=1.0\linewidth]{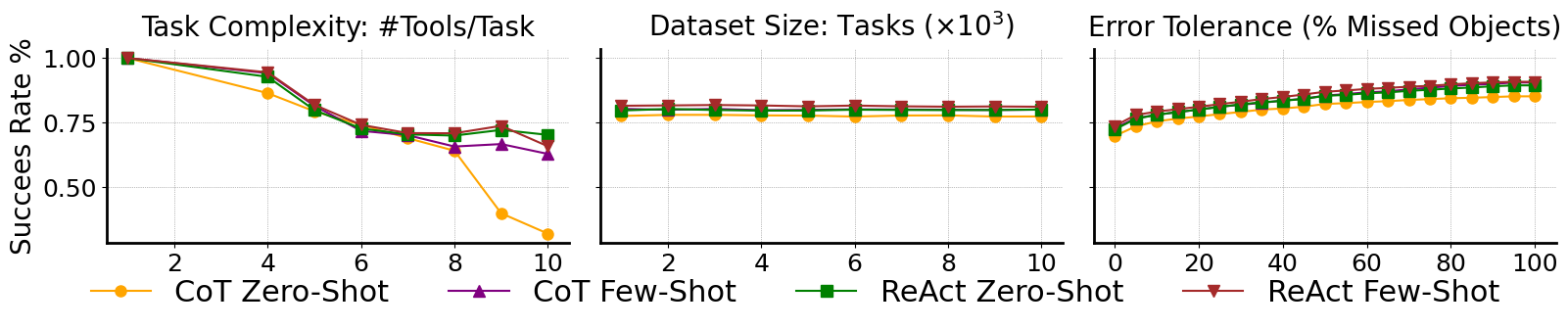}
\caption{Success Rate vs. Task Complexity (left), Benchmark size (middle), and Error Tolerance (right). All methods use GPT-4-Turbo.}
\label{fig:success_rates}
\end{figure*}

\begin{figure}[t!]
\centering
\includegraphics[width=1.0\linewidth]{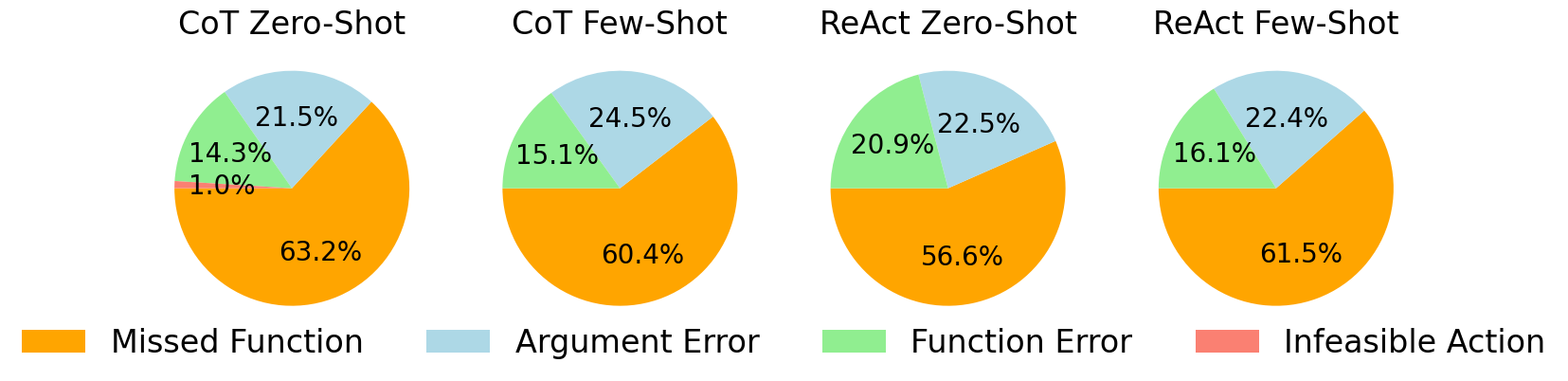}
\caption{Distribution of different errors in correctness rate for CoT and ReAct methods. All methods use GPT-4}
\label{fig:error_types}
\end{figure}

\textbf{Prompting Schemes:} We observe that ReAct generally outperforms CoT for both GPT versions, as confirmed across several domains deploying ReAct for multimodal tasks~\cite{yang2023mmreact}. Moreover, Chameleon's performance is notably lower than that of CoT and ReAct, especially evident with the GPT-4 model, indicating that while Chameleon is a valuable method, its prompting ``decomposition'' strategy which is primarily multi-step single-tool prompting makes it less suited for long-horizon (multi/single-step multi-tool) geospatial tasks. Please note that our focus is tool-augmented fine-tuning-free agents, so we leave the exploration of fine-tuned LLMs for future work.

\textbf{Model Cost:} An interesting observation is the lack of a clear correlation between the tokens consumed and the success or correctness rate. Despite Chameleon consuming the highest number of tokens on average, it does not lead in terms of success or correctness rates. This finding suggests token usage does not directly translate to higher performance, challenging the assumption that more extensive responses might yield better results. These insights collectively highlight the nuanced dynamics of model performance within the \texttt{GeoLLM-Engine} benchmark.

\textbf{Success Rate vs. Task Complexity:} 
Delving into the relationship between success rate (SR) and task complexity provides further insights into agent performance. As shown in \cref{fig:success_rates}, there is an inverse relationship between SR and the number of tool calls required to answer a prompt. Specifically, note how for tasks requiring a single tool call, SR exceeds 95\%, while for more complex tasks involving more than eight tool calls SR is below 70\% and even dropping below 27\% for certain agents. This finding highlights a \textbf{critical limitation} of current geospatial agents: while agents can handle simpler tasks with relative ease, their performance degrades as the task complexity increases. We emphasize the need for benchmarks that measure agents' real-world utility against complex geospatial scenarios.

\textbf{Success Rate vs. Benchmark size:}
Next, we conduct ablations varying dataset size from 500 to 10,000 tasks (\cref{fig:success_rates}): for all methods, the success rates remain relatively unchanged across this range. In a significant escalation of our testing regime, we scaled our \texttt{GeoLLM-Engine} benchmark to 100,000 queries covering over half a million tool calls (\cref{tab:results_scale}). For this experiment, which is the largest in this domain to our knowledge, we leverage the massively parallel nature of our engine over 100 GPT endpoints. Despite the tenfold increase, the success rate of GPT-4 CoT showed relative stability, suggesting that merely increasing the benchmark size does not necessarily challenge the agents more or provide a better assessment of their capabilities. Instead, it is the task complexity within the benchmarks that presents a significant factor in evaluating geospatial performance. These findings critique the prevalent approach in recent works where the focus has been on scaling the benchmark size, while our findings highlight that increasing the complexity of tasks is more essential for assessing agent performance.

\begin{table}[t!]
  \centering
  \resizebox{\columnwidth}{!}{%
  \begin{tabular}{@{}lcccc@{}}
    \toprule
    \multirow{2}{*}{Benchmark} & Number & Number & Success \\
     & of Queries & of Tasks & Rate\\
    \midrule
    \texttt{GeoLLM-Engine-10k} & 10,000 & 50,830 & 77.35\% \\
    \texttt{GeoLLM-Engine-100k} & 100,000 & \textbf{521,868} &  76.81\%  \\
    \bottomrule
  \end{tabular}
  }
  \caption{Success Rate of GPT-4 CoT on massively large-scale benchmark with 100k queries}
  \label{tab:results_scale}
\end{table}

\textbf{Success Rate vs. Error tolerance:}
Last, we capture \texttt{GeoLLM-Engine}'s parameterizable nature that allows us to evaluate agents based on varying error thresholds by properly updating the $\epsilon$ in the model-checkers functions. The incremental increase in success rates with the relaxation of error thresholds reflect scenarios allowing for a certain margin of error or applications where absolute precision is often unattainable (\textit{e.g.}, nuanced definition of what a user means when asking ``\textit{Show me all car detections in Madrid}''; whether this assumes to include suburban areas or not might vary per user).

\begin{table*}[ht!]
  \centering
  \resizebox{0.95\textwidth}{!}{%
  \begin{tabular}{@{}lccccccc@{}}
    \toprule
    Geospatial & Dynamic & Standalone & Diverse &  Functional  & Number & Number & Avg. Query \\
    Benchmarks & Agent? &  UI Engine?  & Commands? &  Correctness?  & of Images & of Queries & Complexity \\
    \midrule
    Charting~\cite{roberts2024charting} & \cmark &\cmark & \xmark & \xmark & $\sim$ 1,000 & - & 1 task/query\\
    GeoChat~\cite{kuckreja2023geochat}  & \cmark &\cmark & \cmark & \xmark & $\sim$ 150,000 & 318,000 & $\sim$1 task/query\red{$^\dag$}\\
    ChatEarthNet~\cite{yuan2024chatearthnet} &\xmark &\xmark & \cmark & \xmark & 173,488 & 173,488 & $\sim$1 task/query\red{$^\dag$}\\
    GRAFT~\cite{mall2023remote} &\xmark &\xmark & \cmark & \xmark & 18,9000,000 & - & 1 task/query\\
    RSVG~\cite{zhan2023rsvg} & \xmark &\xmark & \cmark & \xmark & 17,402 & 38,320 & 1 task/query \\
    RSGPT~\cite{hu2023rsgpt} & \xmark &\xmark & \xmark & \xmark & - & 2,585 & 1 task/query\\
    Remote-CLIP~\cite{liu2024remoteclip} & \xmark &\xmark & \cmark & \xmark & - & - & 1 task/query\\
    RS-CLIP~\cite{silva2024large} & \xmark &\xmark & \xmark & \xmark & 45,134 & 225,670 & 1 task/query\\
    RSICD~\cite{lu2018rsicd} & \xmark &\xmark & \cmark & \xmark & 10,921 & 24,333 & 1 task/query\\
    RS-ChatGPT~\cite{guo2024remotesensinggpt} & \cmark &\cmark & \cmark & \xmark & - & - & $\sim$1 task/query \red{$^\dag$}\\
    SkyEyeGPT~\cite{zhan2024skyeyegpt} & \cmark &\cmark & \cmark & \xmark & - & 968,000 & $\sim$1 task/query \red{$^\dag$}\\
    SkyScript~\cite{wang2023skyscript} & \xmark &\cmark & \cmark & \xmark & 2,600,000 & 1,300,000 & 1 task/query\\
    RS5M~\cite{zhang2024rs5m} & \xmark &\xmark & \cmark & \xmark & $>$5,000,000 & - & 1 task/query\\
    EarthGPT~\cite{zhang2024earthgpt} & \cmark &\cmark & \cmark & \xmark & - & $>$1,000,000 & $\sim$1 task/query \red{$^\dag$}\\
    RSVQA~\cite{lobry2020rsvqa} & \xmark &\xmark & \cmark & \xmark &  1,066,316 & 10,659 & 1 task/query \\
    LHRS-Bot~\cite{muhtar2024lhrsbot} & \xmark &\xmark & \cmark & \xmark & 15,000 & 1,150,000 & $\sim$1 task/query\red{$^\dag$}\\
    \midrule
    \texttt{GeoLLM-Engine} & \cmark & \cmark & \cmark & \cmark & 1,149,612 & 521,868 & 5.23 tasks/query\\
    \bottomrule
  \end{tabular}
  }
  \caption{Comparison \texttt{GeoLLM-Engine} to existing geospatial LLM-related benchmarks. \red{$^\dag$}Note: while these works consider dynamic chatGPT-like prompting capabilities, the underlying agent execution is conducted mainly across multiple steps via conversational prompting of separate tasks (\textit{i.e.}, \textit{multi-step single-task}), rather that multiple tasks (tools) per query (\textit{i.e.}, \textit{single/multi-step \textbf{multi}-task}).}
  \label{tab:sota}
\end{table*}

\textbf{Correctness Rate:} 
\cref{fig:error_types} shows the error types for CoT and ReAct on GPT-4, in both zero-shot and few-shot scenarios. The most common error type is ``Missed Function'', suggesting that the GPT-4 often omits necessary tool calls regardless of the approach used, accounting for more than half of all errors. These confirm recent findings that dynamic/RAG-augmented~\cite{srinivasan2023nexusraven} prompting further improves agent performance by addressing such failures. Last, the consistent error distribution across different methods implies that these issues are not method-specific but rather inherent to the current capabilities of the underlying GPT-4.

\section{Related Work}
\label{sec:related_work}

Recent advancements in autonomous agents span from reinforcement learning platforms \cite{DBLP:journals/corr/BrockmanCPSSTZ16} to web-based applications and benchmarks for web interactions \cite{pmlr-v70-shi17a, DBLP:journals/corr/abs-1802-08802, deng2023mind2web, NEURIPS2022_82ad13ec, zhou2023webarena}. VisualWebArena \cite{koh2024visualwebarena} represents a milestone of LLM benchmarking, offering realistic tasks to evaluate multimodal web agents towards improving performance on complex web pages. The adoption of multimodal models for EO tasks is gaining momentum (\cref{tab:sota}). Innovations like SkyEyeGPT \cite{zhan2024skyeyegpt} and Remote Sensing ChatGPT \cite{guo2024remotesensinggpt} showcase advancements in integrating VQA agents and computer vision models with RS imagery for enhanced multimodal responses. However, existing benchmarks often rely on predefined, single-step text-image prompts. \texttt{GeoLLM-Engine} allows us to assess agents' ability to execute nuanced EO tasks.
\section{Limitations and Future Work}
\label{sec:limitations}

We recognize limitations within our framework. First, using GPT-4 for both generating and evaluating ground truths could introduce bias risks, as highlighted by benchmarking work~\cite{zhou2023webarena}. We are currently enhancing sampling diversity by leveraging hybrid strategies that incorporate both GPT-generated outputs and programmatic elements~\cite{zhuang2023toolqa}. Second, while our platform emulates complex tasks reflective of EO analysts' workflows, we emphasized depth (\textit{long-horizon} tasks) over breadth in task complexity. We are actively expanding the capabilities of our engine to incorporate a wider variety of tasks (\textit{e.g.}, from maritime traffic analysis~\cite{bastani2023satlaspretrain} to illegal fishing~\cite{agnew2009estimating} to damage assessment~\cite{robinson2023rapid}) leveraging \texttt{GeoLLM-Engine}'s flexible APIs. Furthermore, we have focused on assessing finetuning-\textit{free} agents, which is why \texttt{GeoLLM-Engine} doesn't specify train-val-test splits. Following recent work on tuning LLMs on EO tasks~\cite{zhan2024skyeyegpt}, we are expanding our benchmark to tool-agents training~\cite{jian2023stable}. Last, we have utilized standard LLM APIs without explicitly optimizing for cost (\textit{e.g.}, latency or token usage). We are currently orthogonal optimizations to enhance \texttt{GeoLLM-Engine} efficiency, such as LLM compilers~\cite{kim2024llm}, dynamic tooling~\cite{fore2024geckopt}, and token compression~\cite{jiang2023llmlingua}.
\section{Conclusion}
\label{sec:conclusion}

In this work, we introduce \texttt{GeoLLM-Engine}, a novel environment for evaluating geospatial Copilots, designed to bridge the gap between simplistic benchmarks and the complex demands of Earth Observation (EO) applications. By leveraging a rich array of geospatial API tools, dynamic interfaces, and a massive parallel processing framework over 100 GPT-4-Turbo nodes, our environment facilitates the execution of over half a million multifaceted tasks across 1.1 million satellite images. This advancement not only highlights the limitations of existing benchmarks but also sets a new standard for the development and evaluation of AI agents in the geospatial domain. Looking forward, \texttt{GeoLLM-Engine} paves the way for future research to explore sophisticated EO tasks, promising significant strides toward realizing the full potential of geospatial Copilots.
{
    \small
    \bibliographystyle{ieeenat_fullname}
    \bibliography{main}
}

\end{document}